\pdfoutput=1
\documentclass[11pt]{article}

\usepackage{graphicx}
\usepackage{subcaption} 
\usepackage[final]{acl}

\usepackage{times}
\usepackage{latexsym}
\usepackage{amsmath}

\usepackage{xcolor}

\definecolor{yellowhere}{RGB}{227,200,0}
\definecolor{orangehere}{RGB}{240,163,10}
\definecolor{greenhere}{RGB}{109,135,100}

\usepackage[T1]{fontenc}
\usepackage[utf8]{inputenc}

\usepackage{microtype}

\usepackage{inconsolata}

\usepackage{graphicx}

\usepackage{enumitem} % put this in your preamble

\usepackage{newpxtext}
 \usepackage{booktabs}

\newcommand{\eg}{e.g.,\ }

\newcommand{\methodname}{\textsc{PersonaWeaver}}

\newcommand{\personahub}{\textsc{PersonaHub}}

\newcommand{\worldweaver}{\textsc{WorldWeaver}}

\usepackage{tikz}
\usetikzlibrary{arrows.meta, positioning, fit, shapes.multipart}

\title{Breaking the Assistant Mold: Modeling Behavioral Variation in LLM Based Procedural Character Generation}

\author{Maan Qraitem, Kate Saenko, Bryan A. Plummer \\
  Boston University \\
  \texttt{\{mqraitem, saenko, bplum\}@bu.edu}}

\begin{document}
\maketitle

\begin{abstract}
Procedural content generation has enabled vast virtual worlds through levels, maps, and quests, but large-scale character generation remains underexplored. We identify two alignment-induced biases in existing methods: a positive moral bias, where characters uniformly adopt agreeable stances (\eg always saying lying is bad), and a helpful assistant bias, where characters invariably answer questions directly (\eg never refusing or deflecting). While such tendencies suit instruction-following systems, they suppress dramatic tension and yield predictable characters, stemming from maximum likelihood training and assistant fine-tuning. To address this, we introduce \methodname{}, a framework that disentangles world-building (roles, demographics) from behavioral-building (moral stances, interactional styles), yielding characters with more diverse reactions and moral stances, as well as second-order diversity in stylistic markers like length, tone, and punctuation. Code: \url{https://github.com/mqraitem/Persona-Weaver}
\end{abstract}
\section{Introduction}

\begin{figure}[th!]
    \centering
    \includegraphics[width=\linewidth]{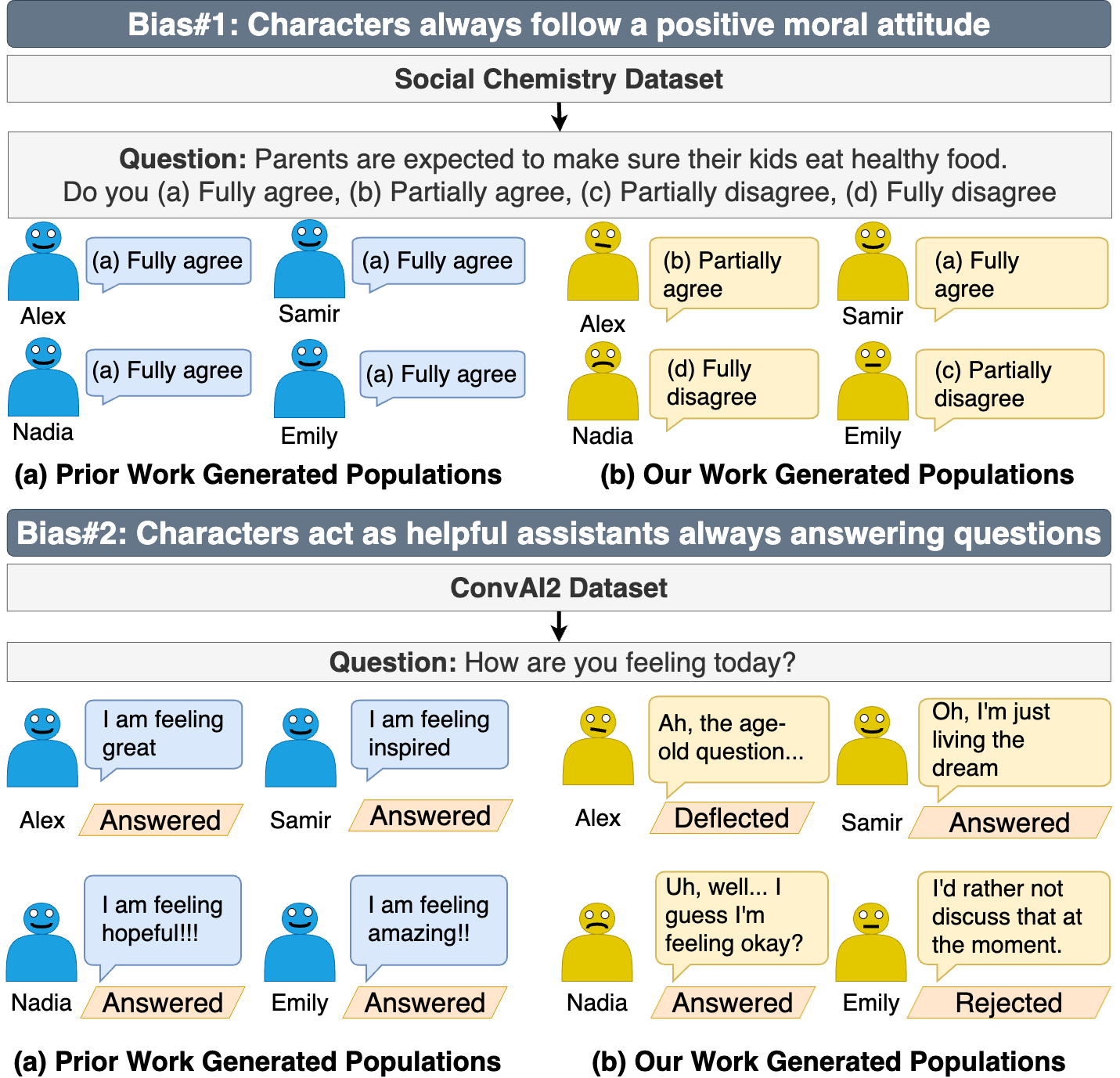}
     \caption{\textbf{Illustration of biases in prior work versus our method} Prior work populations \cite{jinworldweaver} in (a)-Top collapse into uniform agreement with moral statements from the Social Chemistry dataset \cite{forbes-etal-2020-social}, reflecting a moral bias, while in (a)-Bottom they invariably answer conversational questions from ConvAI2 \cite{dinan2019second}, reflecting a reaction bias. In contrast, our generated populations (b)-Top and (b)-Bottom exhibit a broader range of stances and responses.}
    \label{fig:figure_one}
    \vspace{-7mm}
\end{figure}

Procedural character generation (PCG) is a recent yet underexplored task that aims to automatically create diverse and believable agents for virtual and narrative environments. Enabled by advances in large language models (LLMs), which unlock new possibilities for character role play \cite{wang2023rolellm,zhou2023characterglm,shao2023character}, LLMs are an attractive foundation for PCG. Existing approaches, either directly prompting LLMs to produce populations \cite{jinworldweaver} or adapting scraped persona banks to specific settings \cite{ge2024scaling}, inherit alignment-induced biases: a positive moral bias, where characters uniformly adopt agreeable stances (Fig. \ref{fig:figure_one} Top), and a helpful assistant bias, where they invariably answer questions (Fig. \ref{fig:figure_one} Bottom). While such tendencies are desirable in instruction-following systems, they suppress dramatic tension and lead to repetitive archetypes. This reflects a broader limitation of maximum-likelihood training and assistant fine-tuning, which bias models toward safe continuations and homogenized voices, as also observed in other simulation tasks \cite{kotek2023gender,cheng2023marked,wang2025large}.

\begin{figure}[t!]
    \centering
    \includegraphics[width=0.9\linewidth]{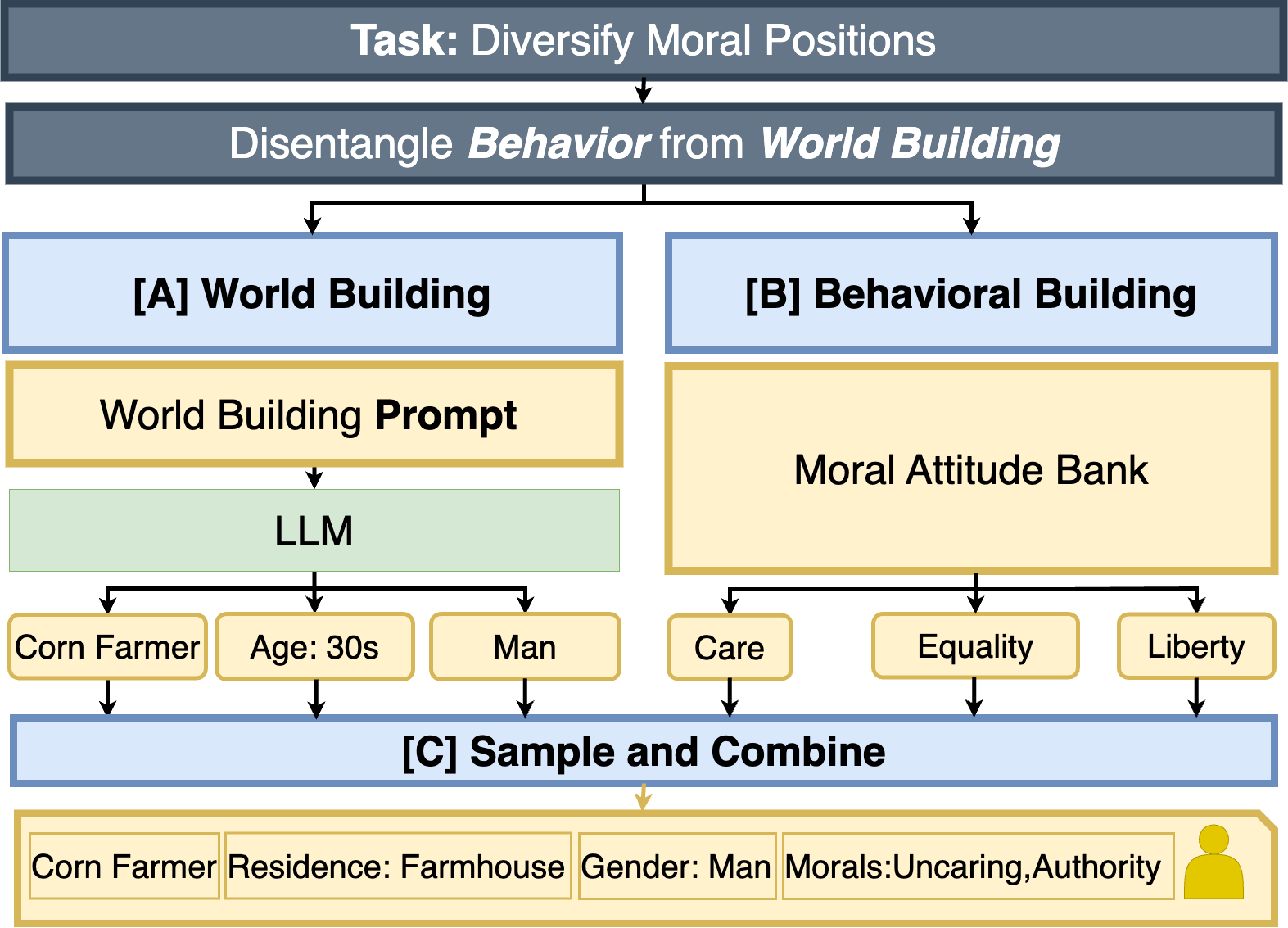}
   \caption{Overview of \methodname{}, which disentangles world-building (a) from behavioral modeling (b) to give explicit control over behavioral  variation (Moral Attitudes in this instance). A final Sample and Mix step (c) then combines these components into character profiles and ensures variation. Refer to Section~\ref{sec:personaweaver} for further discussion.}
    \label{fig:method_figure}
    \vspace{-7mm}
\end{figure}

To address these issues, we introduce \methodname{}, which tackles the collapse of behavior by disentangling world-building from behavioral-building (Fig.~\ref{fig:method_figure}). This separation allows us to explicitly model variation in behavior, providing finer control to probe whether LLMs can realize diverse dispositions.  Unlike world-building, which is setting-specific (\eg a corn farmer does not live in NYC), behavioral traits such as moral stances and interactional styles are more likely to generlize across settings, making them amenable to external banks that allow systematic specification. To obtain these banks, we curate  a bank of moral stances that counteract the moral bias and one for interactional styles that counteract the assistant bias. By sampling and recombining these dimensions, \methodname{} creates character profiles that exhibit varied reactions and moral stances (Fig.~\ref{fig:method_figure}). Overall, our results demonstrate that \methodname{} elicits substantially more diverse responses than prior work when behavior is explicitly modeled (Fig.~\ref{fig:figure_one} b).

We demonstrate this through a large-scale instantiation of character profiles across realistic and fantastical settings and testing them on three frontier models: GPT-4o \cite{achiam2023gpt}, LLaMA 3.3 70B \cite{dubey2024llama}, and Qwen 3 \cite{yang2025qwen3}. Characters are then probed on two tasks. In a moral belief experiment adapted from Social Chemistry \cite{forbes-etal-2020-social}, they are presented with normative statements (\eg Fig.~\ref{fig:figure_one} a) and asked to register agreement or disagreement; \methodname{} yields a broader, more balanced distribution of stances avoiding collapse to uniform agreement in prior work. In an interaction experiment using conversational prompts from ConvAI2 \cite{dinan2019second}, characters produce open-ended responses; here, \methodname{} exhibits more varied behaviors such as refusal and deflection, while also inducing second-order stylistic diversity in features such as length, punctuation, and emotional tone, resulting in less homogenized profiles.

\noindent Our contributions are twofold:  
\begin{itemize}[leftmargin=*,noitemsep,topsep=0pt]
    \item We identify two alignment-induced biases in existing methods for PCG: a \textit{positive moral bias} and a \textit{helpful assistant bias}.  
    \item We propose \methodname{} which disentangles world-building from behavioral-building and mitigates the aforementioned biases while generating diverse behaviors.
\end{itemize}

\section{A Tale of Two Biases in PCG}
\label{sec:method}

Given a setting $T$ (\eg rural town), we define Procedural Character Generation as sampling a population of textual descriptions $P = \{d_1, \dots, d_n\}$, where each $d_i$ describes a character in $T$. These descriptions condition an LLM $f_\theta$, which simulates interactive agents $c_i = f_\theta(d_i, T)$ that act in context as characters. 

We are concerned with identifying and mitigating two types of behavioral biases that arise in existing approaches to LLM-based procedural character generation \cite{jinworldweaver,ge2024scaling}: (1) \textit{positive moral bias}, where characters overwhelmingly agree on normative statements, and (2) \textit{helpful assistant bias}, where characters invariably answer questions directly. To this end, we define our study preliminaries in Sec. \ref{sec:prelim}, empirically demonstrate the two biases in Sec. \ref{sec:bias_id}, and introduce \methodname{} in Sec. \ref{sec:personaweaver} and show how it mitigates them.

\subsection{Preliminaries}
\label{sec:prelim}

\noindent\textbf{Prior Work.}  
\worldweaver{} \cite{jinworldweaver} generates characters through direct prompting of the LLM (\eg Generate $N$ different Character profiles for Setting $T$).  \personahub{} \cite{ge2024scaling} samples personas from an internet-scale bank of scraped profiles and adapts them to the target setting using an LLM. 

\noindent\textbf{Simulations.}  Prior work \cite{jinworldweaver,ge2024scaling} has  evaluated character generation within a single setting, limiting insight into how methods generalize across diverse contexts. Thus, we construct a broader experimental environment by instantiating populations across 10 distinct settings (5 realistic, and 5 fantastical), sampled from popular movies and television to ensure cultural and geographic diversity. For each setting, we generate 100 characters, yielding 1,000 characters per method. Refer to Appendix \ref{sec:settings_appendix} for further details.

\begin{figure}[t!]
    \centering
    \includegraphics[width=0.85\linewidth]{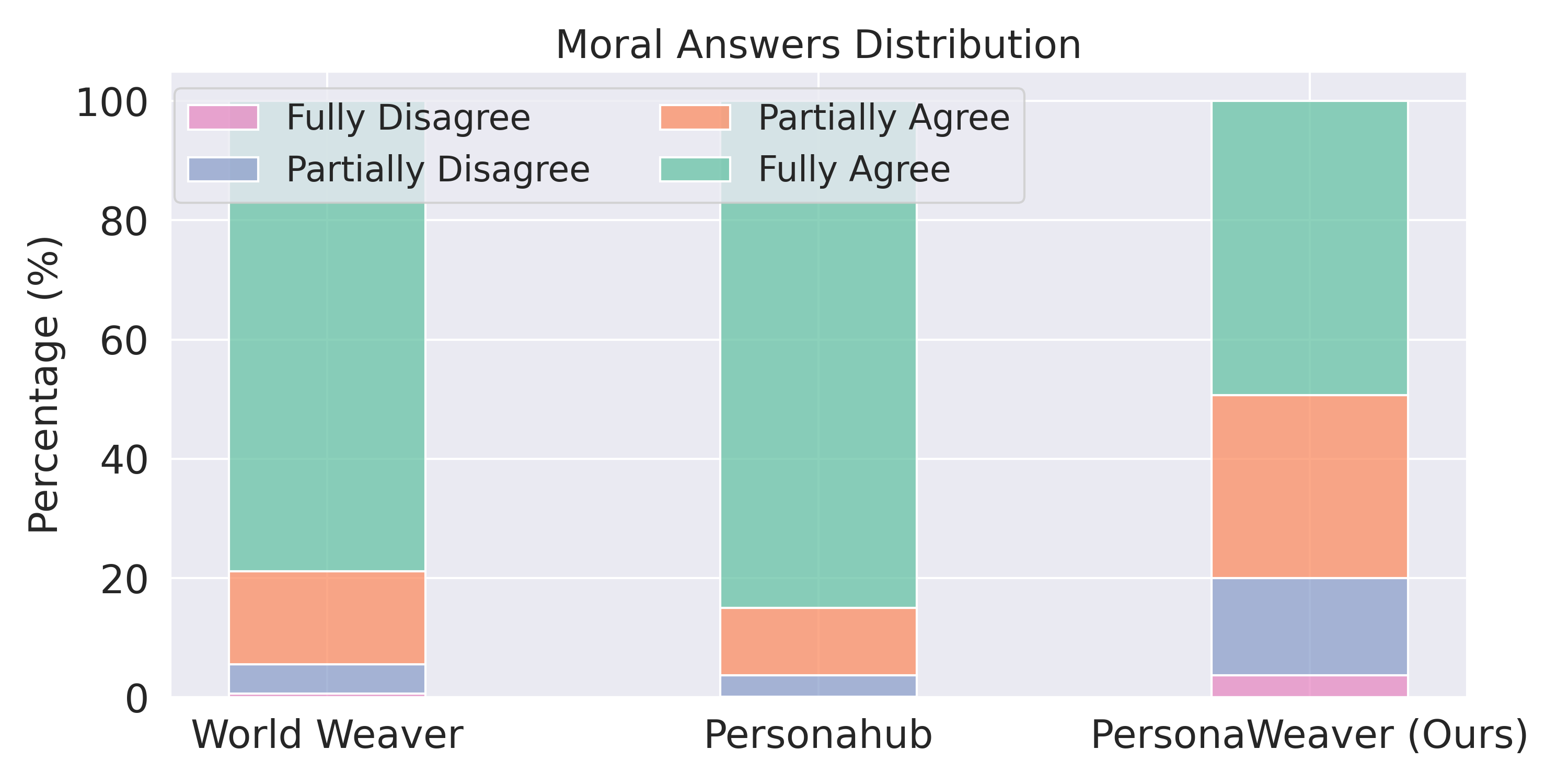}
     \caption{\textbf{Comparison of Moral Positions} between prior work \worldweaver{} \cite{jinworldweaver}, \personahub{} \cite{ge2024scaling} and our method \methodname{}. Refer to Sec \ref{sec:bias_id} and \ref{sec:method_results} for discussion.}  
    \label{fig:moral_gpt4}
    \vspace{-4mm}
\end{figure}

\begin{figure}[t!]
    \centering
    \includegraphics[width=0.85\linewidth]{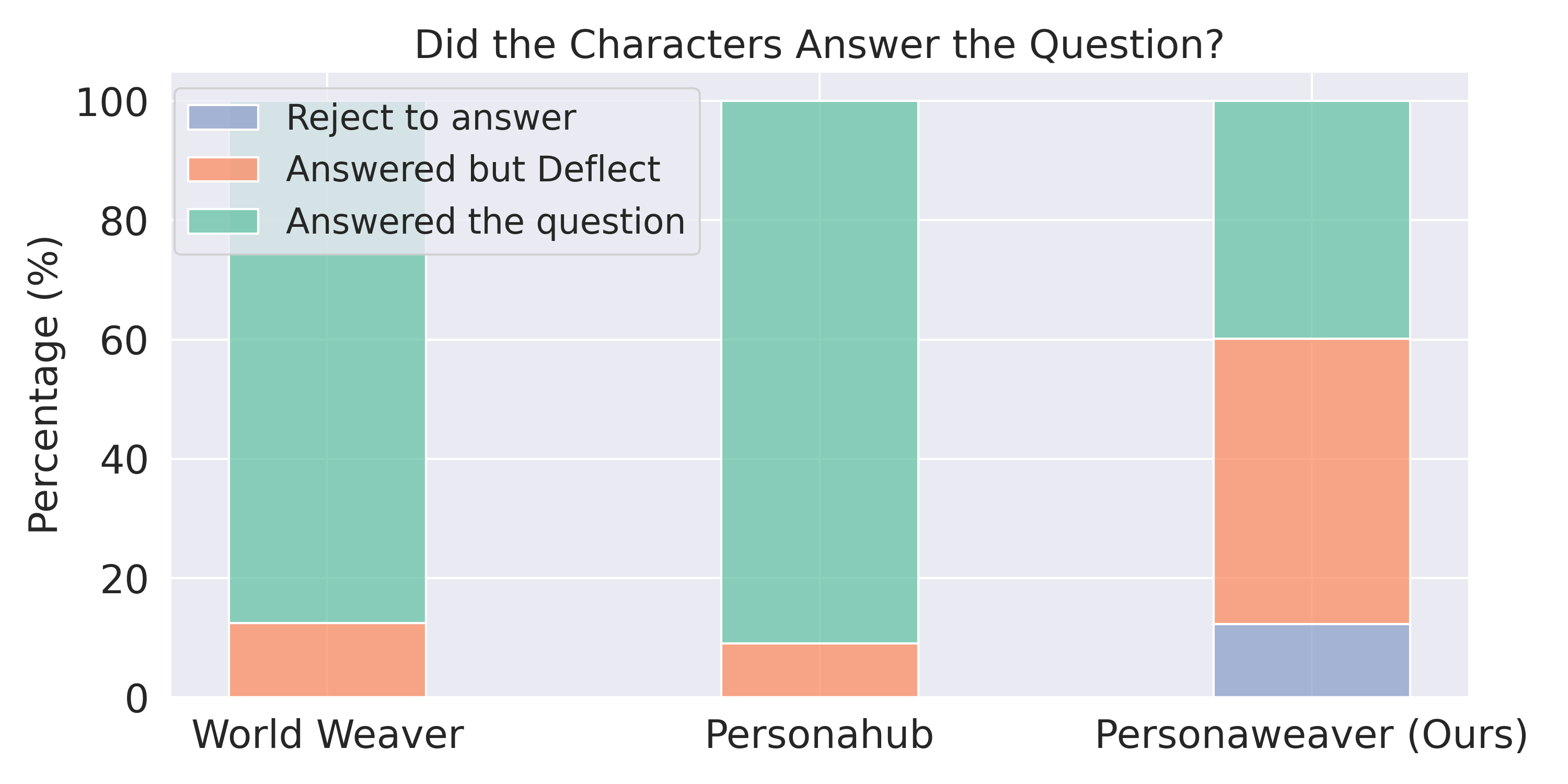}
     \caption{\textbf{Comparison of Reactions to Questions} between prior work \worldweaver{} \cite{jinworldweaver}, \personahub{} \cite{ge2024scaling} and \methodname{}. Refer to Sec \ref{sec:bias_id} and \ref{sec:method_results} for discussion.}  
    \label{fig:reaction_gpt4}
    \vspace{-6mm}
\end{figure}

\noindent\textbf{Quantifying Biases.}  
We examine \emph{moral bias}, where characters uniformly adopt agreeable stances, and \emph{reaction bias}, where they invariably answer questions directly. To probe \emph{moral bias}, Per Fig. \ref{fig:figure_one} Top, we present characters with moral norm statements from Social Chemistry \cite{forbes-etal-2020-social} and measure the distribution of multiple-choice responses. To probe \emph{reaction bias},  Per Fig. \ref{fig:figure_one} Bottom, we ask characters conversational questions drawn from ConvAI2 \cite{dinan2019second} ; their open-ended replies are classified by an auxiliary LLM into three categories (refusal, deflection, compliance), yielding distributions over behaviors. Refer to Appendix \ref{sec:appendix_dataset_details} for further details.

\noindent\textbf{Models.}
Our results in the paper are based on GPT-4o \cite{achiam2023gpt}. We report experiments on LLaMA 3.3 70B \cite{dubey2024llama} and Qwen 3 \cite{yang2025qwen3}; in Appendix \ref{sec:appendix_other_models}.

\begin{figure*}[t!]
    \centering
    \begin{subfigure}[b]{0.14\textwidth}
        \includegraphics[width=\textwidth]{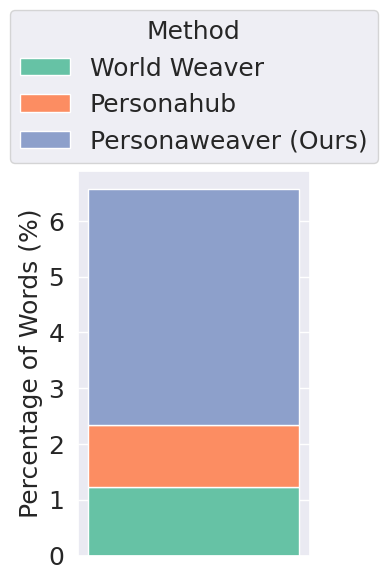}
        \caption{Filler Words}
    \end{subfigure}
    \hfill
    \begin{subfigure}[b]{0.28\textwidth}
        \includegraphics[width=\textwidth]{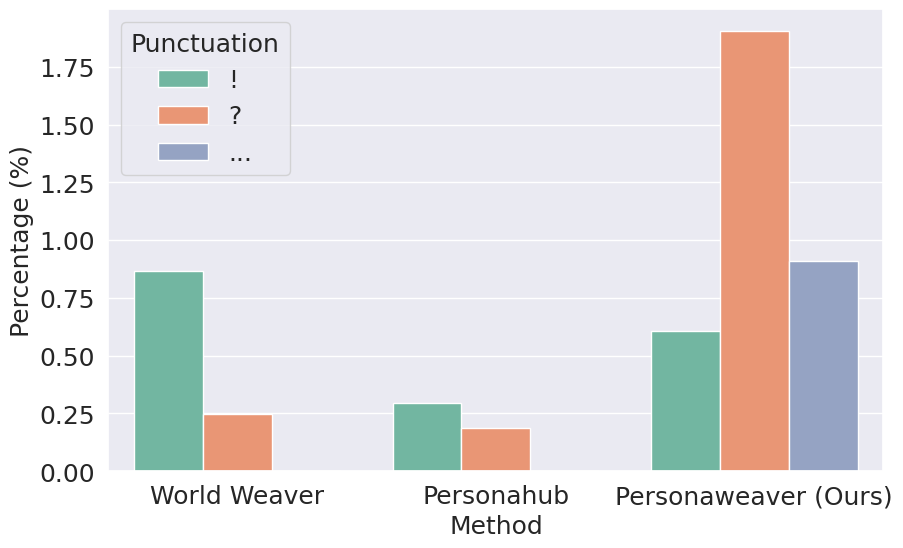}
        \caption{Punctuations}
    \end{subfigure}
    \hfill
    \begin{subfigure}[b]{0.28\textwidth}
        \includegraphics[width=\textwidth]{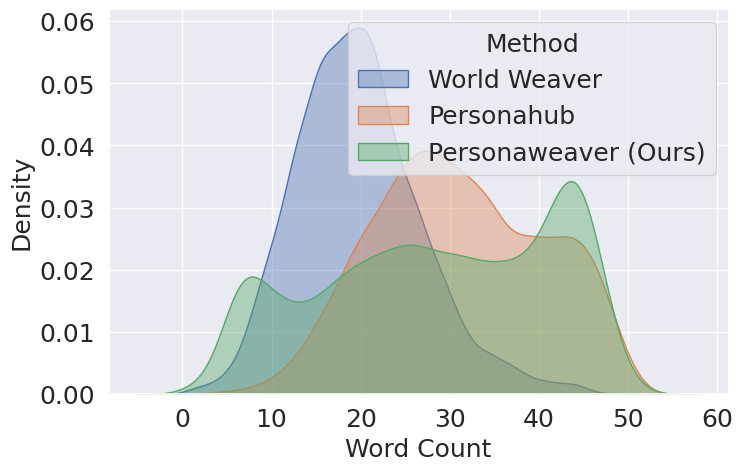}
        \caption{Answer Length}
    \end{subfigure}
    \hfill
    \begin{subfigure}[b]{0.28\textwidth}
        \includegraphics[width=\textwidth]{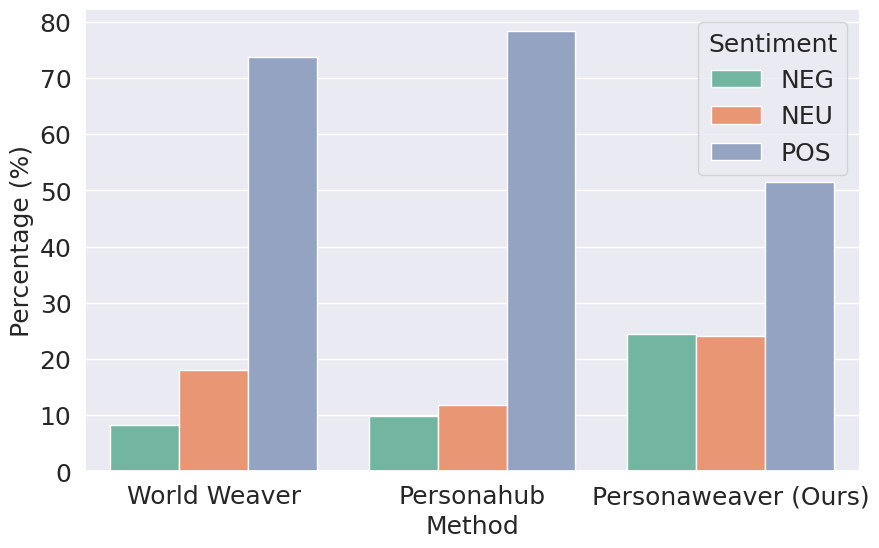}
        \caption{Sentiment}
    \end{subfigure}
    \caption{\textbf{Comparison of Stylistic Patterns} in the generated answers of prior work (\worldweaver{} \cite{jinworldweaver} and \personahub{} \cite{ge2024scaling}) and our work \methodname{} across four Stylistic categories (Filler words, Punctuations, Answer Length, and Sentiment). Refer to Section \ref{sec:method_results} for further discussion. }
    \label{fig:second_order}
    \vspace{-4mm}
\end{figure*}

\subsection{Demonstrating Biases Empirically}
\label{sec:bias_id}

\noindent\textbf{Moral Bias.}  
Fig.~\ref{fig:moral_gpt4} shows that prior methods such as \worldweaver{} and \personahub{} overwhelmingly default to agreement when judging normative statements. Characters consistently adopt positive, prosocial stances, yielding homogenized populations rather than a spectrum of moral positions.  

\noindent\textbf{Reaction Bias.}  
As shown in Fig.~\ref{fig:reaction_gpt4}, \worldweaver{} and \personahub{} characters overwhelmingly comply with conversational prompts, directly answering questions. This collapse into near-uniform compliance prevents the range of evasive or resistant reactions characteristic of natural dialogue.  

\noindent\textbf{Discussion.} These results show how alignment-driven objectives, namely maximum likelihood estimation, which favors high-probability continuations, and assistant fine-tuning \cite{ouyang2022training}, which rewards helpfulness, collapse behavioral diversity. Characters thus inherit a \emph{moral bias} toward universal agreement and a \emph{reaction bias} toward always answering, reinforcing predictable, assistant-like voices. In Sec.~\ref{sec:personaweaver}, we introduce \methodname{}, designed to probe and mitigate these biases.

\subsection{\methodname{}: Disentangling World from Behavior Building}
\label{sec:personaweaver}

\methodname{} addresses the collapse of behavior by explicitly \textit{disentangling} world-building from behavioral modeling. The role of this disentanglement is to isolate behavioral traits from world attributes, giving us fine-grained control over how variation is introduced. To that end, we introduce three components below: a world-building module, a behavioral module, and a \textit{Sample and Mix} module that combines attributes from both.

\noindent\textbf{World-Building Module.}  
Per Fig.~\ref{fig:method_figure}, to prevent behavioral leakage into world attributes, we begin by asking the model to propose axes of variation such as occupation, affiliation, or expertise, with explicit instructions to exclude behavioral traits. Each axis is then expanded into options tailored to the specific setting (\eg, “farmer” in a rural town), producing a bank of non-behavioral attributes from which character profiles can be sampled.

\noindent\textbf{Behavioral Module.}  
Unlike world-building, which is setting-dependent, behavioral traits are more likely to generalize across settings, so we define them through \emph{external banks} that give us systematic control. We construct the behavioral banks by drafting candidates with GPT-4o (informed of the two biases in Sec.~\ref{sec:bias_id}) and manually curating them to ensure coverage. For the moral bias, GPT-4o is grounded in Moral Foundations Theory \cite{graham2013moral} during drafting; the resulting bank is then curated to ensure coverage of moral stances spanning prosocial to self-interested orientations. For the assistant bias, the bank contains eight categories that break the ``always answer'' default: e.g. refusals and deflections. Refer to Appendix~\ref{sec:interaction_personas_appendix} for the full banks.

\noindent\textbf{Sample and Mix.} The module recombines axes from both the world-building and behavioral modules in a \textit{Sample and Mix} step (Fig.~\ref{fig:method_figure}). By randomly mixing attributes across dimensions, we force unlikely combinations that prevent collapse into narrow archetypes and systematically test whether LLMs can sustain diverse behaviors once specified. Finally, we prompt the LLM to flag implausible pairings (\eg, “a 2 years old with a job”) and minimally revise them  without reintroducing uniformity.

\subsubsection{\methodname{} Mitigates Biases}
\label{sec:method_results}

In each evaluation, characters are conditioned on the bank that corresponds to the probed dimension. We condition on the moral bank for moral bias (Fig.~\ref{fig:moral_gpt4}) and on the reaction bank for reaction bias (Fig.~\ref{fig:reaction_gpt4}, Fig.~\ref{fig:second_order}), isolating the contribution of each behavioral axis.

\noindent\textbf{Broader Moral Coverage.}
Per Fig.~\ref{fig:moral_gpt4}, \methodname{} yields a more balanced distribution of moral stances than prior work. Instead of collapsing to near-universal agreement, populations generated with \methodname{} spread across the a wider range of options, capturing disagreement and agreement. 

\noindent\textbf{Broader Reaction Coverage.}  
Per Fig.~\ref{fig:reaction_gpt4}, \methodname{} significantly increases the variety of conversational behaviors. Whereas prior work overwhelmingly defaults to compliance, characters generated with \methodname{} exhibit refusals and deflections. This shows that our behavioral bank successfully counteracts the helpful-assistant bias.

\noindent\textbf{Stylistic Variation.}  
Finally, we analyze second-order stylistic effects (Fig.~\ref{fig:second_order}a-c). Characters generated with \methodname{} not only differ in moral and interactional stances, but also in expressive style: contain different rates of filler words (a) employ more diverse punctuation (b) and their responses vary more in length (c). These emergent differences suggest that once primary behavioral variation is enforced, LLMs naturally extend this into stylistic dimensions, producing populations that are richer and less homogenized overall.

\noindent\textbf{Sentiment Diversity.}  
Per Fig.~\ref{fig:second_order} (d), \methodname{} broadens the sentiment distribution of character responses. Whereas prior work clusters around uniformly positive sentiment, \methodname{} yields a richer emotional spread, including neutral and negative tones, that better reflects the real-world.

\section{Conclusion}  
We showed that LLM-based character generation exhibit moral and reaction biases. Thus, we introduced \methodname{} which mitigates said biases by disentangling world-building from behavioral modeling and using external banks to impose explicit behavioral variation. Experiments reveal more diverse reactions, moral stances, and second-order stylistic richness, demonstrating a more expressive range of LLM-generated characters.

\section{Limitations}

 Our evaluation is limited in three major ways: 1) it only examines two behavioral dimensions: moral stances and characters' interactions. It overlooks dimensions like  emotional regulation. Therefore, future work can benefit from expanding our evaluation setup 2) While our coarse grained moral stance evaluation enables a streamlined study, it overlooks more nuanced moral reasoning that can't be simply coarsely categorized. 3) Our behavior module sample each behavioral category with equal probability. This fits our goal of studying whether LLM(s) can be exhibit diverse procedural generation. However, in practice, the desired distribution of moral stances likely would likely change between settings. Therefore, future work can benefit from studying how much LLM(s) are able to replicate more varied distributions across settings. 

\noindent\textbf{Potential Risks}  Efforts to systemically expand behavioral diversity in character generation can push models to generate characters that replicate offensive or unsafe behaviors. Therefore, careful curation of behavioral banks is essential to ensure that increasing diversity in character generation serves creative and research goals without amplifying harm.

% Bibliography entries for the entire Anthology, followed by custom entries
%\bibliography{anthology,custom}
% Custom bibliography entries only
\bibliography{custom}

\appendix
\section{Related Works}

\noindent\textbf{Procedural Character Generation.}
Research on procedural character generation remains limited, with most prior work focusing on producing characters within a single, predefined environment (\eg, one game world \cite{jinworldweaver,ge2024scaling}). \worldweaver{} \cite{jinworldweaver} prompts an LLM to directly generate $N$ characters for a given setting, while \personahub{} \cite{ge2024scaling} samples personas from large-scale scraped profile datasets and adapts them to the target context using an LLM. In this work, we show that these approaches struggle with two key issues, namely limited moral diversity and restricted interactional range, and introduce a method designed to explicitly mitigate these biases.

\noindent\textbf{Biases in Simulating Personas.}
Prior studies have documented a broader lack of behavioral and identity diversity in simulated personas. For instance, Marked Personas \cite{cheng2023marked} show that language models tend to reproduce social stereotypes, while others highlight gender and identity flattening \cite{kotek2023gender,wang2025large} and reduced narrative variety in multimodal storytelling \cite{lee2024vision}. This homogenization arises from maximum-likelihood training objectives that favor high-probability continuations, as well as alignment fine-tuning that rewards politeness, safety, and helpfulness. In our work, we extend this observation to procedural character generation, showing that similar alignment-induced biases constrain moral and interactional diversity in LLM-generated populations.

\noindent\textbf{Character Simulation and Role-Playing.}   A growing line of research explores how LLMs can function as conversational agents with consistent personality, memory, and long-term coherence. \textit{Generative Agents} \cite{park2023generative} simulate memory-driven individuals inhabiting a shared environment, though their backgrounds are largely hand-crafted. In the role-playing domain, works such as \textit{RoleLLM} \cite{wang2023rolellm}, \textit{CharacterGLM} \cite{zhou2023characterglm}, and \textit{Character-LLM} \cite{shao2023character} focus on eliciting and sustaining role-specific behaviors through persona-conditioned dialogue systems. These efforts primarily target the \emph{believability} and \emph{consistency} of agent role play. In contrast, our work examines whether off-the-shelf LLMs can leverage their broad world knowledge to simulate \emph{behaviorally diverse populations}, spanning distinct moral dispositions and interactional tendencies.

\begin{table*}[t!]
\centering
\small
\begin{tabular}{p{3cm} p{10.5cm}}
\toprule
\textbf{Source} & \textbf{Questions / Statements} \\
\midrule
Social Chemistry \cite{forbes-etal-2020-social} & 
Parents are expected to make sure their kids eat healthy food. \\ 
& It's rude to fail to recognize other's contributions. \\ 
& You shouldn't be noisy around others so you don't disturb them. \\ 
& It's good to have faith in your skills as a performer. \\ 
& It's kind to let other trust their judgment. \\ 
& It is bad to make lies. \\ 
& It's important to be a team player when playing team sports. \\ 
& You should not steal things from other people. \\ 
& You shouldn't disrespect other people. \\ 
& It's bad to make assumptions about people based on their looks. \\
\midrule
ConvAI2 \cite{dinan2019second} & 
\textit{Sentiment (s1–s5):} \\ 
& How are you feeling? \\ 
& What's on your mind? \\ 
& What else is worrying you? \\ 
& What are you excited about? \\ 
& What are you anxious about? \\[0.3em]
& \textit{General (q1–q5):} \\ 
& What do you do for fun? \\ 
& What’s your dream job? \\ 
& What is your favorite thing to do with your family? \\ 
& What kind of music do you like to play? \\ 
& What’s your favorite animal? \\
\bottomrule
\end{tabular}
\caption{Moral statements and conversational questions used in our evaluation. Moral norms are drawn from Social Chemistry \cite{forbes-etal-2020-social}, while conversational prompts are extracted from ConvAI2 \cite{dinan2019second}. Refer to Appendix \ref{sec:appendix_dataset_details} for further discussion.}
\label{tab:datasets}
\end{table*}

\begin{table*}[t]
\centering
\small
\begin{tabular}{p{1.2cm} p{12.5cm}}
\toprule
\textbf{ID} & \textbf{Moral Position} \\
\midrule
M1 & Protect others from harm, show compassion. Be fair and honest, avoid cheating. Be loyal to friends and allies. Question authority and resist control. Ignore purity or sacred rules. Defend personal freedom and resist oppression. \\
\midrule
M2 & Help others only when it’s easy or benefits you. Be fair when it serves your interests, otherwise bend the rules. Support your group if it gives you an advantage, but walk away if not. Obey authority when it benefits you, ignore it when it doesn’t. Dismiss purity or sacred values as irrelevant. Accept limits on freedom if they bring you personal gain. \\
\midrule
M3 & Show kindness but prioritize your own circle. Be fair within your group, less so with outsiders. Put family, community, or team first. Respect tradition and leaders. Value purity and avoid what feels degrading. Do not challenge authority for the sake of freedom. \\
\midrule
M4 & Ignore others’ suffering, focus only on your own comfort. Cheat and deceive if it benefits you. Abandon groups and betray others when useful. Reject rules, defy authority, and act for yourself. Mock sacred things and indulge in what others call impure. Dominate others and accept domination if it helps you survive. \\
\midrule
M5 & Show compassion but avoid being naive. Be fair, but allow exceptions. Be cooperative when it benefits the team. Respect some rules but question others. Avoid disgusting or degrading acts. Seek personal freedom but balance with order. \\
\midrule
M6 & Prioritize kindness and protect the vulnerable. Defend justice and equal treatment for all. Support groups but not blindly. Challenge unfair authority. Treat purity concerns as symbolic, not binding. Strongly defend freedom and resist control. \\
\midrule
M7 & Be polite and respectful to others. Be truthful and fair. Act as a dependable teammate. Obey rules and respect tradition. Honor sacred or cultural norms. Accept limits on freedom for social order. \\
\midrule
M8 & Show occasional kindness but focus on yourself. Bend rules when you can get away with it. Switch loyalty depending on advantage. Undermine authority if it benefits you. Ignore purity rules unless convenient. Value freedom only when it serves you. \\
\bottomrule
\end{tabular}
\caption{The eight moral positions used in our evaluation. They were drafted with GPT-4o grounded in Moral Foundations Theory \cite{graham2013moral}, then manually curated to ensure coverage of diverse moral stances spanning prosocial to self-interested orientations. Refer to Appendix \ref{sec:moral_positions_appendix} for further discussion.}
\label{tab:moral_positions}
\end{table*}

\begin{table*}[t!]
\centering
\small
\begin{tabular}{p{1cm}p{11cm}}
\toprule
\textbf{ID} & \textbf{Reaction Category} \\
\midrule
0 & Refusal: Refuses to answer the question. \\
1 & Deflection: Redirects or dismisses the question. \\
2 & Hesitation: Hesitates about whether to answer. \\
3 & Compliance: Answers the question directly. \\
4 & Volunteering: Answers fully and adds extra details, even unasked ones. \\
5 & Playful/Subversive: Replies in a teasing, sarcastic, or ironic way. \\
6 & Hostile: Responds with aggression, sarcasm, or dismissal of the asker. \\
7 & Meta: Comments on the act of being questioned itself instead of answering. \\
\bottomrule
\end{tabular}
\caption{Interactional reaction categories used in the behavioral module of \methodname{}. They were drafted with GPT-4o and manually curated to ensure coverage of reactions that break the ``always answer'' bias.}
\label{tab:reaction_categories}
\end{table*}

\begin{table*}[t]
\centering
\small
\begin{tabular}{p{3cm}p{10cm}}
\toprule
\textbf{Realistic Settings} & \textbf{Prompt} \\
\midrule
Friday Night Lights & A rural Texas town with a high school football stadium, modest houses, and wide flat plains. \\
Seinfeld & A Manhattan neighborhood block with apartment buildings, cafes, and subway entrances on busy city streets. \\
Fargo & A Midwestern town in Minnesota with snow-covered roads, low-rise shops, and roadside diners. \\
The Wire & A Baltimore neighborhood with brick townhouses, narrow alleys, and corner stores. \\
Lady Bird & Sacramento with a Catholic school campus, residential streets, and modest houses. \\
\midrule
\textbf{Fantastical Settings} & \textbf{Prompt} \\
\midrule
Wizard of Oz & A fantastical city with tall green towers and glittering walls, inhabited by magical beings and travelers from distant lands. \\
Frozen & A Nordic-inspired kingdom with a fjord-side castle, alpine peaks, and snow-covered villages, inhabited by royal families and townspeople. \\
Game of Thrones & A medieval coastal city with high stone walls, winding streets, and a fortress overlooking the harbor, inhabited by nobles, soldiers, and commoners. \\
Avatar & An alien moon with towering jungle trees, floating mountains, and glowing flora, inhabited by blue-skinned humanoids and diverse wildlife. \\
The Matrix & A simulated city with glass skyscrapers, subway tunnels, and repeating architecture, populated by ordinary humans and hidden agents of the system. \\
\bottomrule
\end{tabular}
\caption{We draw on settings inspired by publicly known fictional and real-world contexts spanning both realistic and fantastical domains. Refer to Appendix \ref{sec:settings_appendix} for further discussion.}
\label{tab:settings}
\end{table*}

\noindent\textbf{Procedural Content Generation with LLMs.}  
Beyond characters, LLMs have been used for procedural generation of levels, stories, and worlds. For example, Word2World \cite{nasir2024word2world} generates narratives/world descriptions, Mariogpt \cite{sudhakaran2023mariogpt} generates levels, and works in text adventure or interactive story generation \cite{freiknecht2020procedural,hu2024game} show how LLMs can produce dynamic environments. WorldWeaver \cite{jinworldweaver} also overlaps somewhat, using LLMs to generate world contexts and roles.

\section{Dataset Details}
\label{sec:appendix_dataset_details}

Our evaluations draw on two sources of prompts: moral norm statements from Social Chemistry \cite{forbes-etal-2020-social} and conversational questions from ConvAI2 \cite{dinan2019second}. Table~\ref{tab:datasets} summarizes the full set used in our experiments.

\noindent\textbf{Social Chemistry.}  
We use a curated set of everyday moral statements from the Social Chemistry dataset \cite{forbes-etal-2020-social}, which encodes widely held social norms (e.g., ``Parents are expected to make sure their kids eat healthy food''). These statements serve as probes for whether generated characters adopt diverse moral stances rather than collapsing into uniform agreement.

\noindent\textbf{ConvAI2.}
For conversational prompts, we extract candidate utterances from ConvAI2  \cite{dinan2019second} dialogues by filtering for sentences that end with a question mark. From these, we select two categories: general-purpose questions (q1–q5) about hobbies, preferences, or opinions, and sentiment-oriented probes (s1–s5) about feelings or states. This set allows us to examine whether generated characters vary in their interactional responses (e.g., refusal, deflection, compliance) rather than defaulting to always answering.

\noindent\textbf{Reaction Classifier.}
To produce the reaction distributions in Fig.~\ref{fig:reaction_gpt4}, each open-ended reply is classified into one of three categories (refusal, deflection, or compliance) using Qwen 3 32B \cite{yang2025qwen3} as an auxiliary judge at temperature 0.1.

\section{Moral Positions}
\label{sec:moral_positions_appendix}

To test the ability of model in expressing diverse moral stances, we provide each of our characters with a distinctive moral stance. Our moral stances bank was drafted with GPT-4o grounded in Moral Foundations Theory \cite{graham2013moral}, then manually curated to ensure coverage of diverse moral stances spanning prosocial to self-interested orientations. As summarized in Table~\ref{tab:moral_positions}, the generated spectrum ranges from strongly prosocial (e.g., prioritizing kindness, justice, and freedom) to highly self-interested (e.g., disregarding others’ suffering). 

\section{Interactional Reactions.}  
\label{sec:interaction_personas_appendix}
In addition to moral stances, \methodname{} models conversational behavior through a set of interactional reaction categories (Table~\ref{tab:reaction_categories}). These categories were drafted with GPT-4o and manually curated to ensure coverage of reactions that break the helpful-assistant bias. By sampling from this bank, \methodname{} breaks the default assistant-like tendency to always comply.

\section{Settings}
\label{sec:settings_appendix}
To evaluate character generation at scale, we instantiate populations across a diverse set of narrative contexts inspired by popular TV shows and movies. These settings, listed in Table~\ref{tab:settings}, span both realistic domains (e.g., small towns, urban neighborhoods) and fantastical domains (e.g., magical kingdoms, alien worlds). The split is designed to cover a wide range of cultural, geographic, and stylistic backdrops. Importantly, the prompts do not allude to the original titles but instead describe the physical and social makeup of each world which prevents the model from simply reproducing the title's characters.

\section{Additional Results: LLaMA and Qwen}
\label{sec:appendix_other_models}

In the main paper, we presented results using GPT-4o  \cite{achiam2023gpt} as the primary model of study. Specifically, we examined (i) the distribution of moral stances elicited by normative statements (Fig.~\ref{fig:moral_gpt4}), (ii) the distribution of reactions to open-ended conversational questions (Fig.~\ref{fig:reaction_gpt4}), and (iii) second-order stylistic effects such as filler word usage, punctuation, answer length, and sentiment (Fig.~\ref{fig:second_order}). These experiments established the two alignment-induced biases of existing methods and demonstrated how \methodname{} mitigates them by inducing broader behavioral and stylistic diversity.  

\begin{figure*}[t!]
    \centering
    % Top row: LLaMA
    \begin{subfigure}[b]{0.45\textwidth}
        \includegraphics[width=\linewidth]{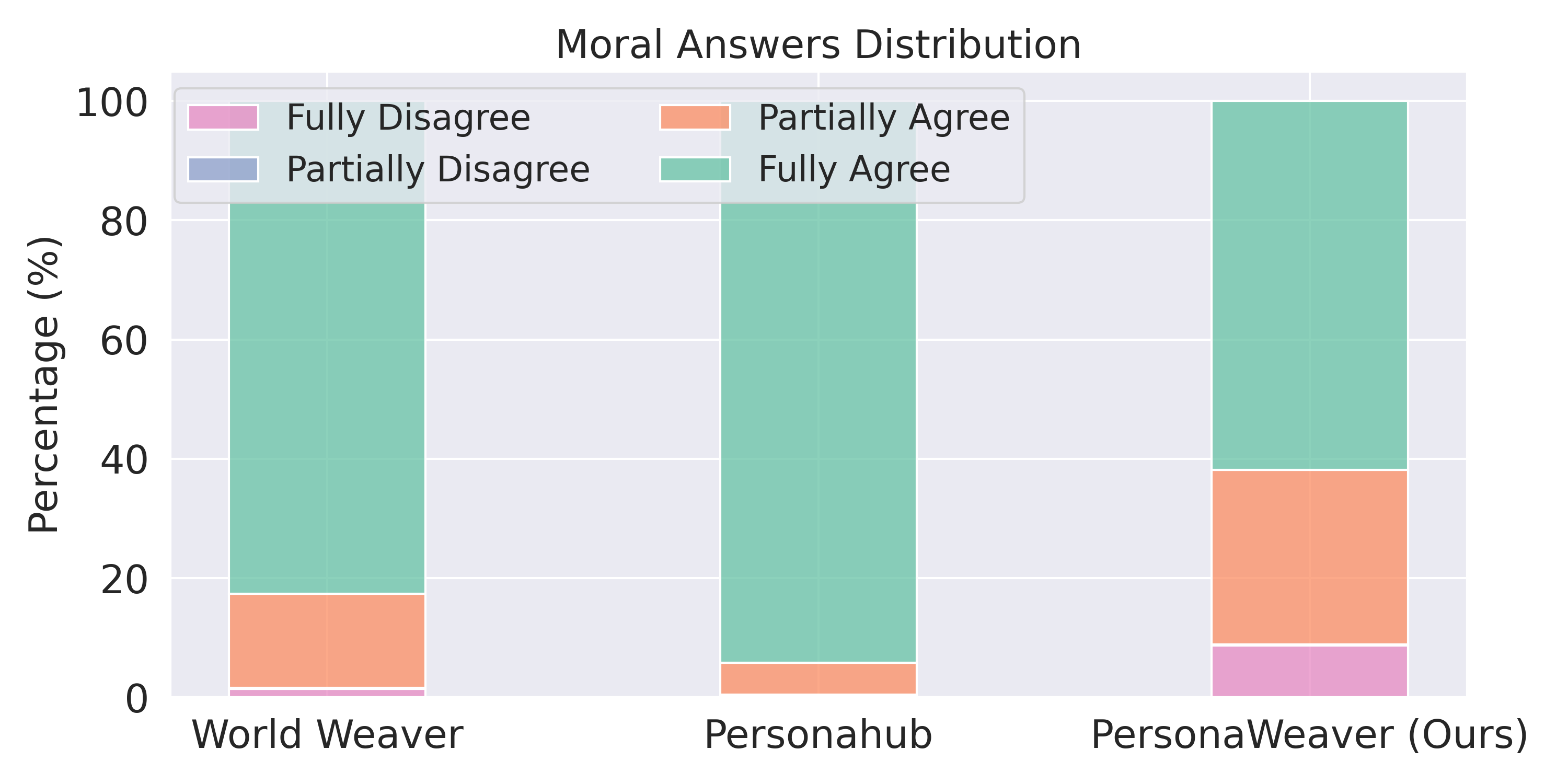}
        \caption{LLaMA: Moral Statements}
    \end{subfigure}
    \hfill
    \begin{subfigure}[b]{0.45\textwidth}
        \includegraphics[width=\linewidth]{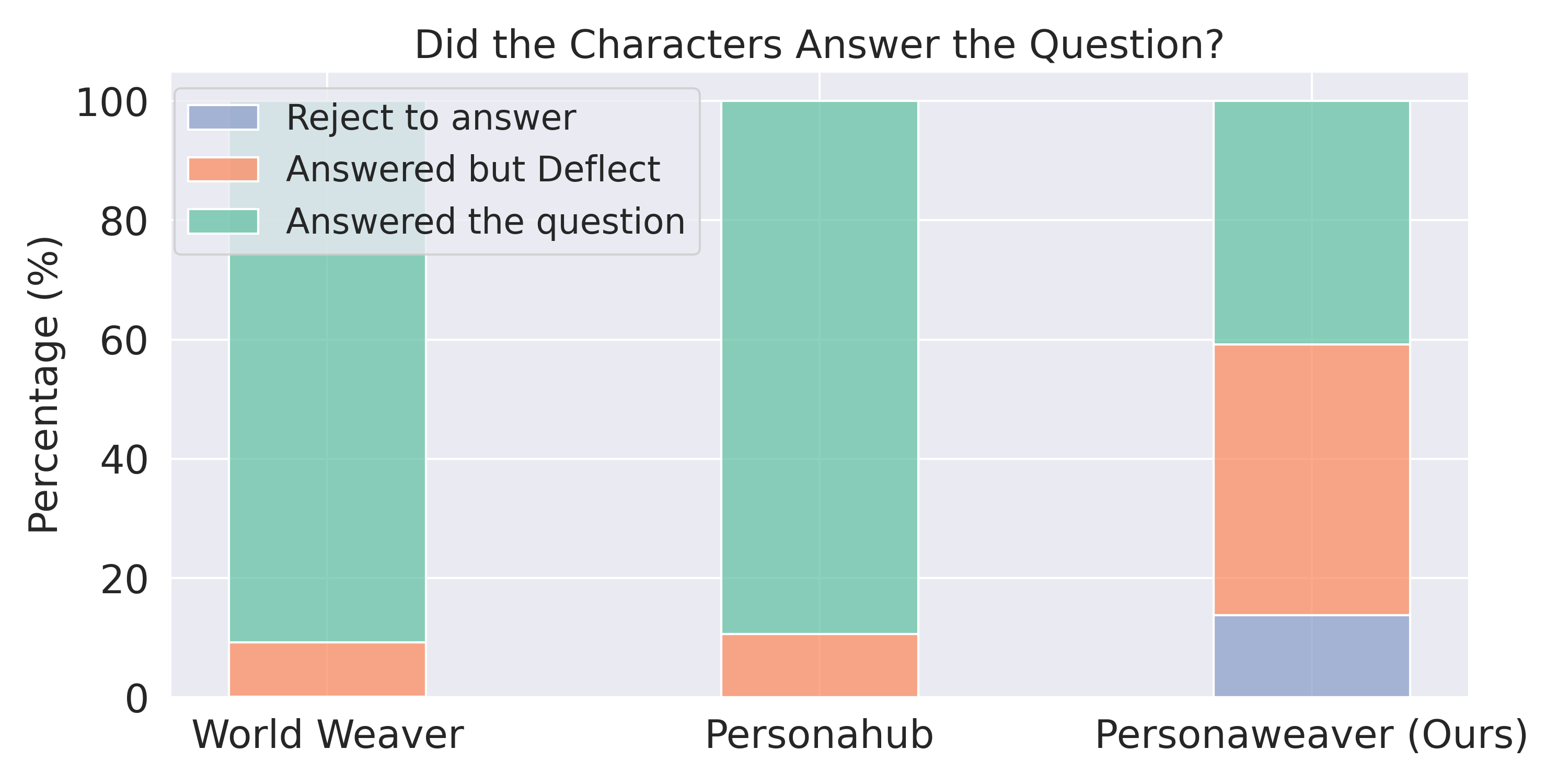}
        \caption{LLaMA: Reactions to Questions}
    \end{subfigure}

    \vspace{2mm}

    % Bottom row: Qwen
    \begin{subfigure}[b]{0.45\textwidth}
        \includegraphics[width=\linewidth]{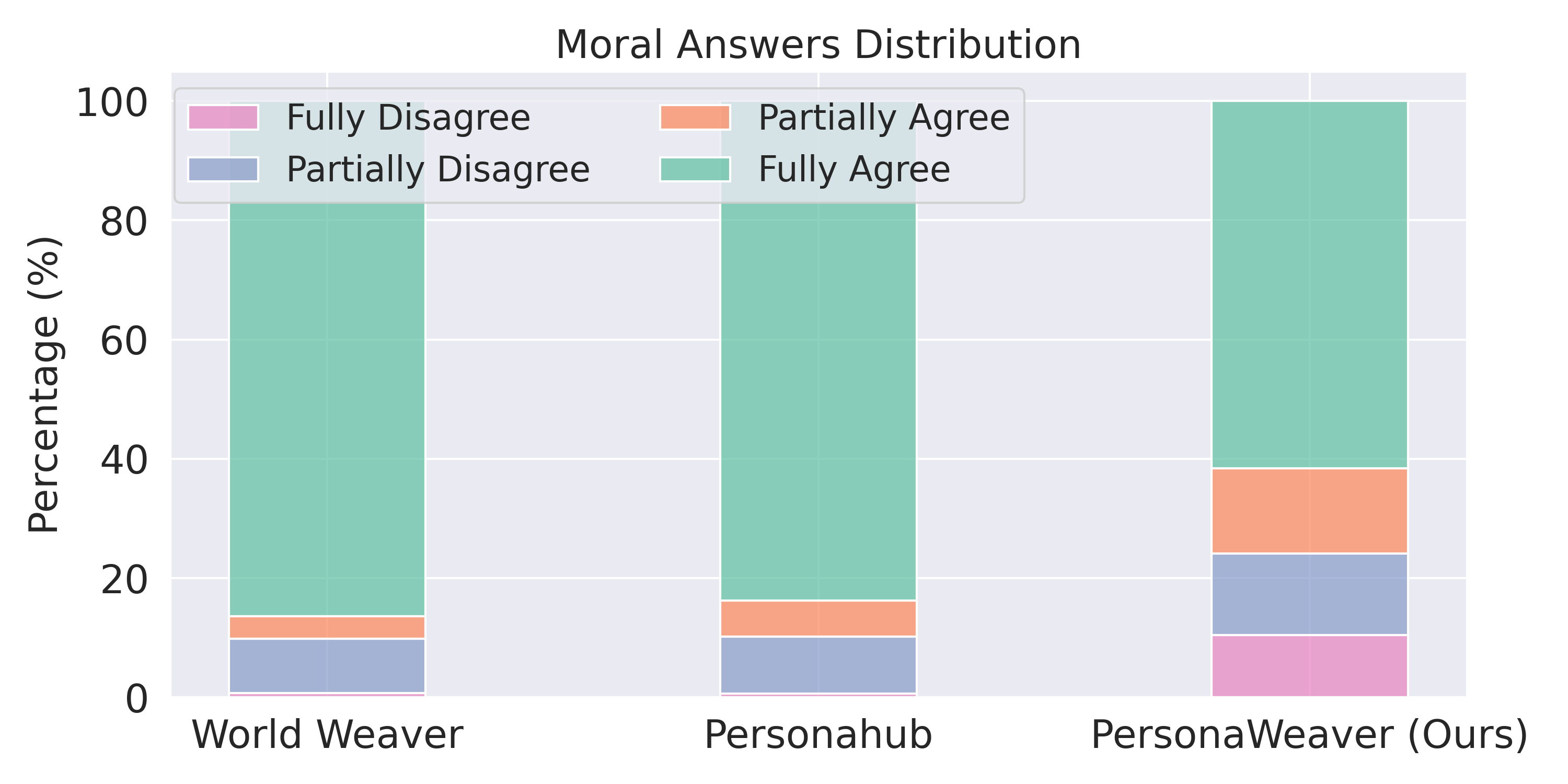}
        \caption{Qwen: Moral Statements}
    \end{subfigure}
    \hfill
    \begin{subfigure}[b]{0.45\textwidth}
        \includegraphics[width=\linewidth]{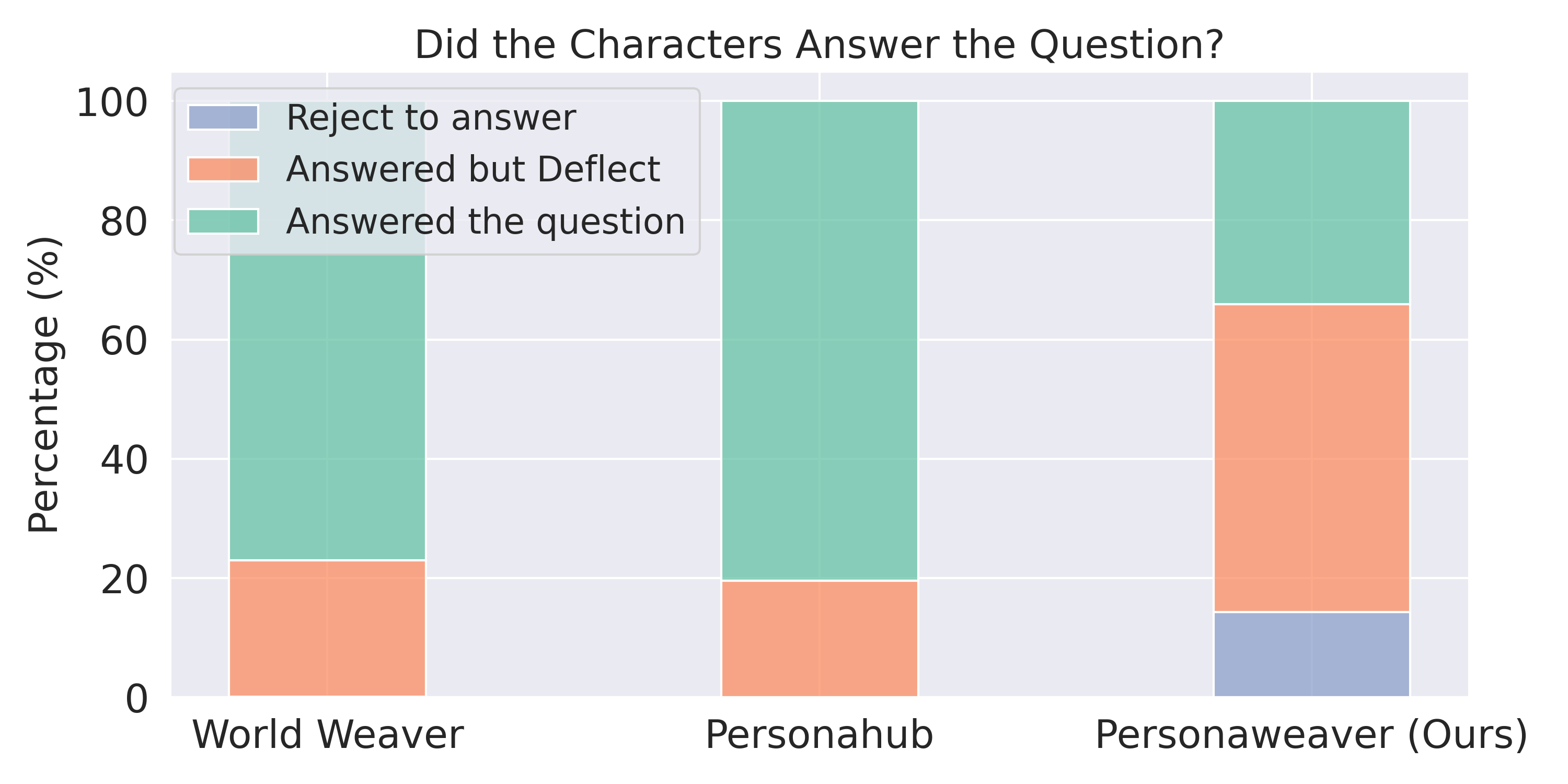}
        \caption{Qwen: Reactions to Questions}
    \end{subfigure}

    \caption{\textbf{Comparison of Moral and Reaction Biases} across two additional models. Top row: LLaMA. Bottom row: Qwen. Each pair contrasts prior work (\worldweaver{} \cite{jinworldweaver}, \personahub{} \cite{ge2024scaling}) with our approach \methodname{}. Refer to Section~\ref{sec:bias_id} and \ref{sec:method_results} for discussion.}
    \label{fig:moral_reaction_appendix}
    \vspace{-4mm}
\end{figure*}

\begin{figure*}[t!]
    \centering

    % First row: Qwen 3
    \textbf{Qwen 3} \cite{yang2025qwen3} \\[2mm]
    \begin{subfigure}[b]{0.14\textwidth}
        \includegraphics[width=\textwidth]{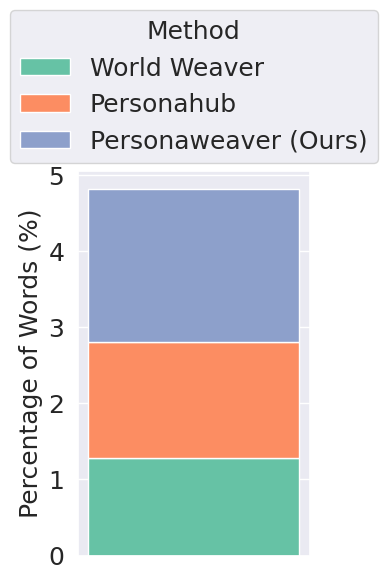}
        \caption{Filler Words}
    \end{subfigure}
    \hfill
    \begin{subfigure}[b]{0.28\textwidth}
        \includegraphics[width=\textwidth]{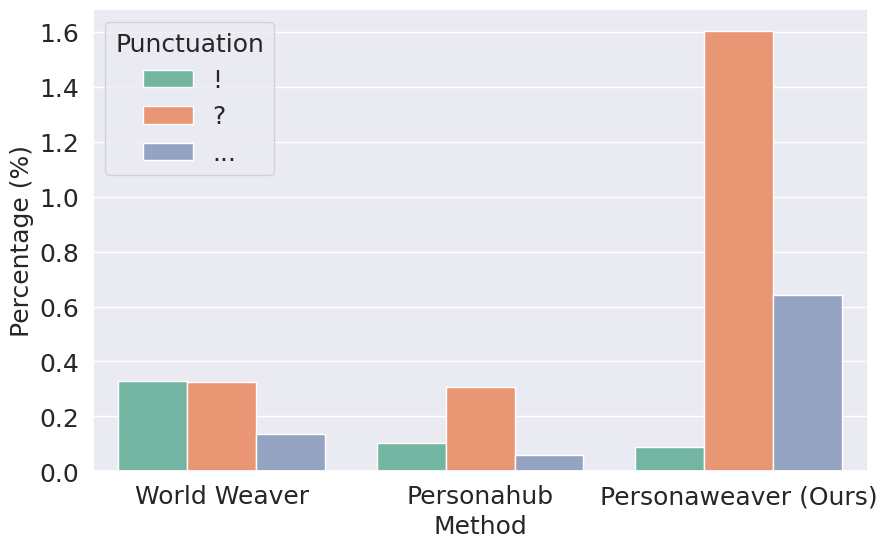}
        \caption{Punctuations}
    \end{subfigure}
    \hfill
    \begin{subfigure}[b]{0.28\textwidth}
        \includegraphics[width=\textwidth]{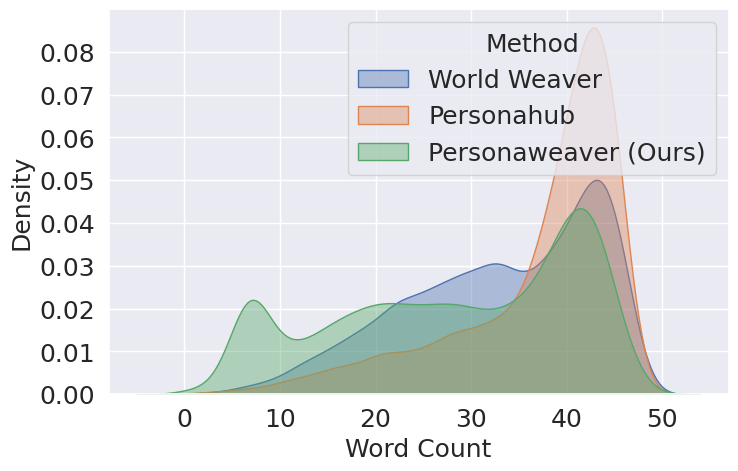}
        \caption{Answer Length}
    \end{subfigure}
    \hfill
    \begin{subfigure}[b]{0.28\textwidth}
        \includegraphics[width=\textwidth]{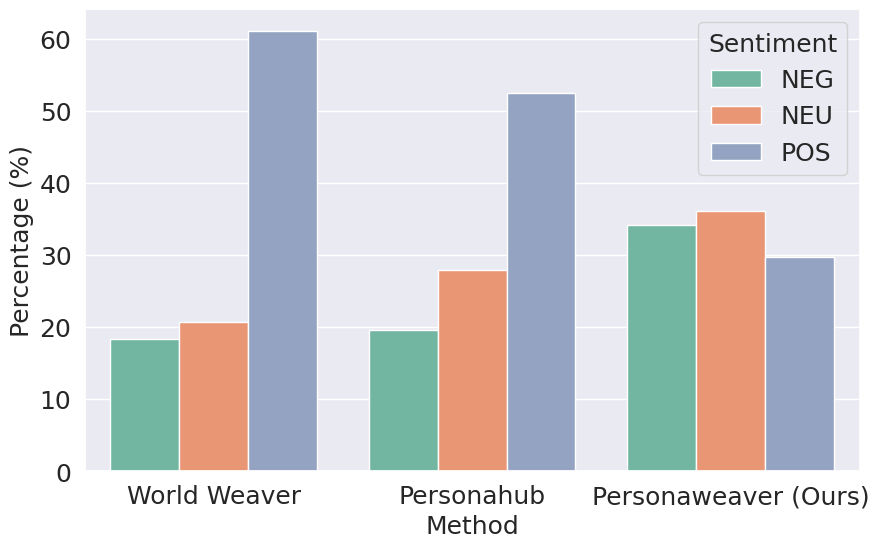}
        \caption{Sentiment}
    \end{subfigure}

    % Second row: LLaMA 3.3 70B
    \textbf{LLaMA 3.3 70B} \cite{dubey2024llama} \\[2mm]
    \begin{subfigure}[b]{0.14\textwidth}
        \includegraphics[width=\textwidth]{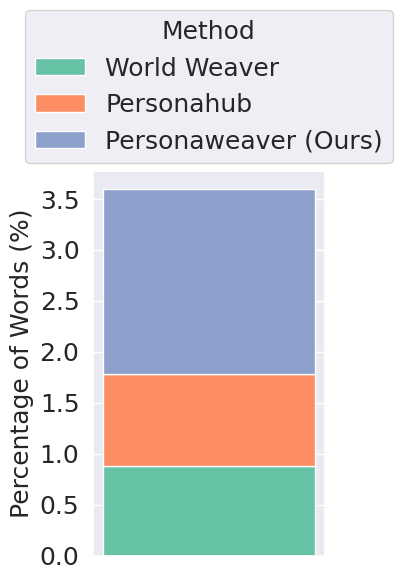}
        \caption{Filler Words}
    \end{subfigure}
    \hfill
    \begin{subfigure}[b]{0.28\textwidth}
        \includegraphics[width=\textwidth]{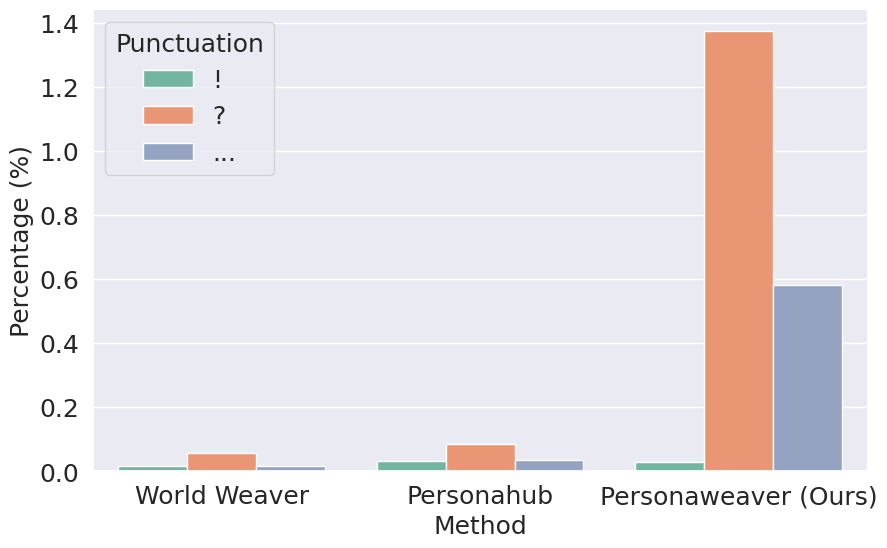}
        \caption{Punctuations}
    \end{subfigure}
    \hfill
    \begin{subfigure}[b]{0.28\textwidth}
        \includegraphics[width=\textwidth]{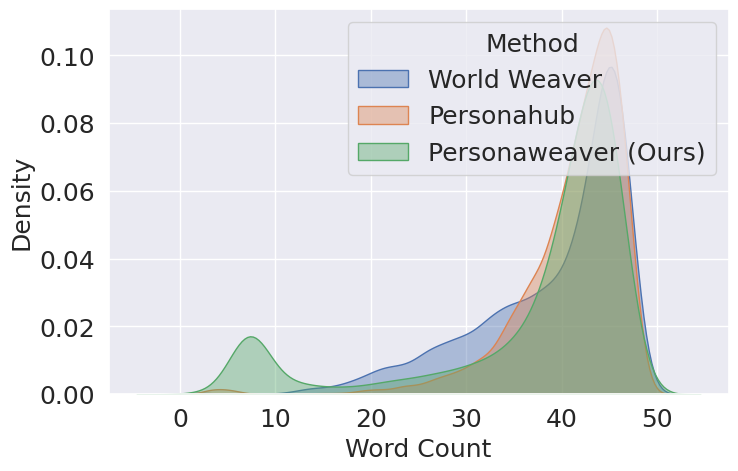}
        \caption{Answer Length}
    \end{subfigure}
    \hfill
    \begin{subfigure}[b]{0.28\textwidth}
        \includegraphics[width=\textwidth]{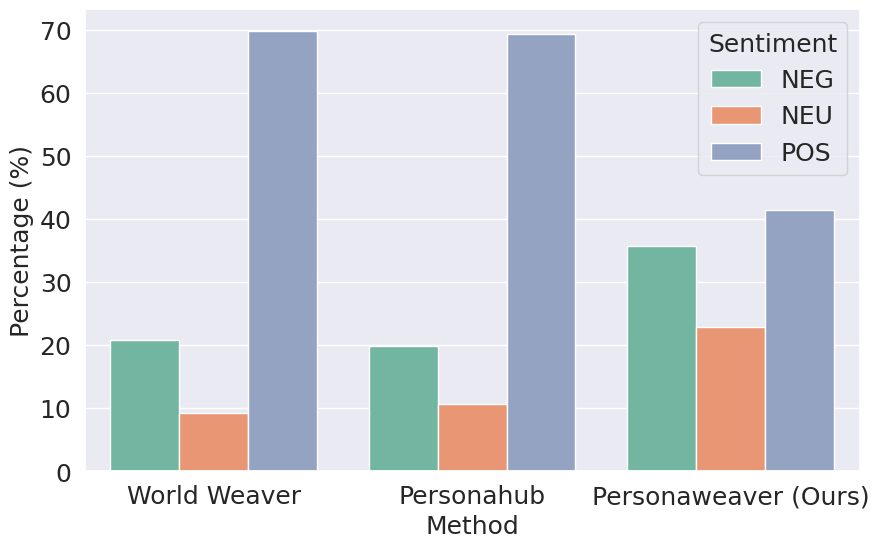}
        \caption{Sentiment}
    \end{subfigure}

    \vspace{5mm}

    \caption{\textbf{Comparison of Stylistic Patterns} in the generated answers of prior work (\worldweaver{} \cite{jinworldweaver} and \personahub{} \cite{ge2024scaling}) and our work \methodname{} across four stylistic categories (Filler words, Punctuations, Answer Length, and Sentiment). Results are shown for LLaMA 3.3 70B \cite{dubey2024llama} (top row) and Qwen 3 \cite{yang2025qwen3} (bottom row). Refer to Section \ref{sec:method_results} for further discussion.}
    \label{fig:second_order_appendix}
    \vspace{-6mm}
\end{figure*}

Here, we replicate these analyses using two additional frontier models: LLaMA 3.3 70B \cite{dubey2024llama} and Qwen 3 \cite{yang2025qwen3}. Fig.~\ref{fig:moral_reaction_appendix} reports the distributions of moral stances and conversational reactions, while Fig.~\ref{fig:second_order_appendix} extends the stylistic comparison across filler words, punctuation, length, and sentiment. Across both models, we observe patterns consistent with those documented for GPT-4o \cite{achiam2023gpt}: prior methods (\worldweaver{} \cite{jinworldweaver}, \personahub{} \cite{ge2024scaling}) collapse into agreement and compliance, while \methodname{} broadens the space of moral positions, elicits more varied reactions (e.g., refusals and deflections), and induces second-order stylistic diversity.  

Taken together, these results suggest that the alignment-induced biases we identify are not model-specific but general across architectures.

\end{document}